\documentclass[journal]{IEEEtran}
\usepackage{cite}
\usepackage{graphicx}
\usepackage{dblfloatfix}  
\usepackage{leftidx}
\usepackage{amsmath,amsfonts,amsthm}
\usepackage{bm} 
\usepackage{booktabs} 
\usepackage[ruled,vlined]{algorithm2e}
\usepackage{caption}
\usepackage[colorlinks,
            linkcolor=black,
            anchorcolor=black,
            citecolor=black
            ]{hyperref}
\begin{document}
\renewcommand*{\figureautorefname}{Fig.}
\captionsetup[figure]{name={Fig.}}

\title{Sequential Decision Fusion for Environmental Classification in Assistive Walking}

\author{Kuangen Zhang, Wen Zhang, Wentao Xiao, Haiyuan Liu, Clarence W. de Silva, and Chenglong Fu
\thanks{K. Zhang, W. Zhang, H. Liu, W. Xiao, and C. Fu are with the Department of Mechanical and Energy Engineering, Southern University of Science and Technology, Shenzhen 518055, China (Corresponding author: Chenglong Fu: fucl@sustc.edu.cn).

K. Zhang and C.W. de Silva are with the Department of Mechanical Engineering, The University of British Columbia, Vancouver V6T1Z4, Canada.}

\thanks{Code and data: \url{https://github.com/KuangenZhang/HMM-decision-fusion} }

\thanks{Digital Object Identifier 10.1109/TNSRE.2019.2935765}

\thanks{\textcopyright 2019 IEEE. Personal use of this material is permitted. Permission from IEEE must be obtained for all other uses, in any current or future media, including reprinting/republishing this material for advertising or promotional purposes, creating new collective works, for resale or redistribution to servers or lists, or reuse of any copyrighted component of this work in other works. }

}




\maketitle
\begin{abstract}
Powered prostheses are effective for helping amputees walk on level ground, but these devices are inconvenient to use in complex environments. Prostheses need to understand the motion intent of amputees to help them walk in complex environments. Recently, researchers have found that they can use vision sensors to classify environments and predict the motion intent of amputees. Previous researchers can classify environments accurately in the offline analysis, but they neglect to decrease the corresponding time delay. To increase the accuracy and decrease the time delay of environmental classification, we propose a new decision fusion method in this paper. We fuse sequential decisions of environmental classification by constructing a hidden Markov model and designing a transition probability matrix. We evaluate our method by inviting able-bodied subjects and amputees to implement indoor and outdoor experiments. Experimental results indicate that our method can classify environments more accurately and with less time delay than previous methods. Besides classifying environments, the proposed decision fusion method may also optimize sequential predictions of the human motion intent in the future.
\end{abstract}
\begin{IEEEkeywords}
Decision fusion, environmental classification, assistive walking, sequential model
\end{IEEEkeywords}

\IEEEpeerreviewmaketitle

\section{Introduction}
\label{sec:Introduction}
\IEEEPARstart{A}{mputation} attenuates the mobility of millions of amputees in daily life. There were 44,430 new lower limb amputees in Canada from 2006 to 2011 \cite{imam_incidence_2017}, and the situation is more serious in the USA. Researchers predicted that the amputee population in the USA would increase to 3.6 million by the year 2050 \cite{ziegler-graham_estimating_2008}. Without healthy lower limbs, these amputees face serious difficulties in daily life. For lower limb amputees, everyday tasks, such as walking and running, present major challenges. In order to help amputees to walk, researchers have developed artificial legs, which are called prostheses \cite{au_powered_2008, au_powered_2009, sup_upslope_2011, lawson_control_2013}. There are two types of prostheses, powered and passive prostheses. Powered prostheses are better than passive prostheses because they can provide the necessary active force to amputees during walking \cite{gailey_energy_1994, wang_walk_2015}. 

Although powered prostheses are efficient for helping amputees walk on level ground, they are inconvenient to use in complex environments. In complex environments, amputees need to switch locomotion modes between different environments (e.g., level ground, up/down stairs, and up/down ramp) \cite{varol_multiclass_2010}, then prostheses should be able to change locomotion modes accordingly. To address this issue, Sup et al. introduced a finite-state controller \cite{sup_design_2008} composed of a series of parametric controllers that uses different parameters in different locomotion modes to control the prosthesis. To achieve seamless switching between different modes, however, the prosthesis must predict the motion intent of the amputee, which is difficult for the prosthesis. Unlike recognizing human activity \cite{wang_deep_2018}, recognizing human intent is difficult to do accurately because the human intent happens mentally and so cannot be measured in the same way.

Previous researchers have primarily focused on the signals in the human-prosthesis loop to predict human motion intent. For instance, targeted muscle reinnervation (TMR) \cite{souza_advances_2014}, electromyography (EMG) \cite{clites_proprioception_2018}, inertial measurement unit (IMU) \cite{xu_real-time_2018}, and mechanical sensors \cite{simon_configuring_2014} have been used to recognize human intent. TMR and EMG allow researchers to measure the electric potential produced by the muscle, which occurs prior to the motion \cite{souza_advances_2014}, but these muscle signals are noisy and are difficult to classify accurately. Signals provided by the IMU and mechanical sensors, on the other hand, are stable but time-delayed \cite{hao_smoother-based_2019}. Moreover, regardless of which signal is used, these signals are user-dependent, which means that they vary for different subjects. Consequently, it is difficult to predict human intent accurately and robustly based on the signals above. 

\begin{figure*}[!b]
    \centering
    \includegraphics[width=16cm]{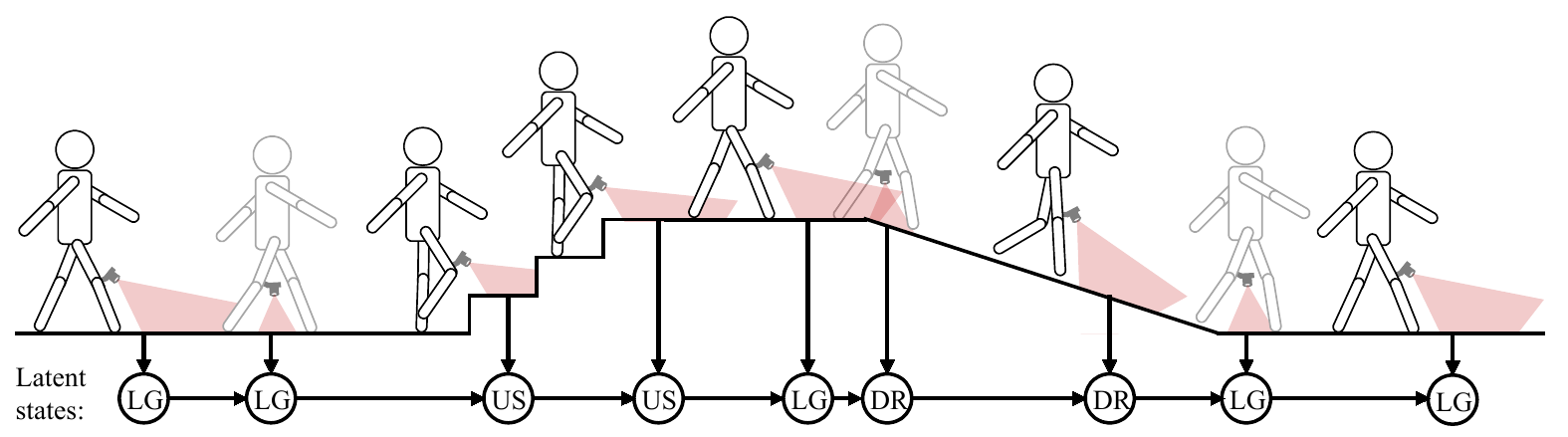}
    \caption{Sequential model of environmental classification. We regard human walking in complex environments as sequential behaviors. To estimate the sequential behaviors, we use the vision system to perceive environments continuously. Current environments are denoted as the latent states, including level ground (LG), up stairs (US), down stairs (DS), up ramp (UP), and down ramp (DR).}
    \label{fig:1-Overview} 
\end{figure*}

Another method to predict the motion intent of amputees is to recognize the signals in the prosthesis-environment loop \cite{zhang_sensor_2019}. Visual information can guide able-bodied people to change locomotion modes in different environments \cite{matthis_gaze_2018}. Similarly, environmental recognition can provide the prosthesis with the environmental context of the human motion intent and help the prosthesis to reconstruct the vision-locomotion loop. The first research to consider combining the visual sensor with the powered prosthesis can be traced back to 2015 when a Kinect camera was used to recognize the geometric parameters of the stairs \cite{krausz_depth_2015}. Subsequently, Liu et al. combined an IMU with a laser sensor to classify five types of terrains, including level ground, up/down stairs, and up/down ramp \cite{liu_development_2016}. Recently, Massalin et al. applied a wearable depth camera to capture the depth images of the environments and designed a support vector machine (SVM) method to classify the environments \cite{massalin_user-independent_2018}. In our previous research \cite{zhang_environmental_2019, zhang_linked_2019}, we utilized a self-contained depth camera and an IMU to capture stable point clouds of environments and designed a graph convolutional neural network (CNN) to classify point clouds. No matter which method is used, original classification results based on the above methods are usually noisy. Researchers have to utilize filters, such as the majority voting filter, to improve classification results. These filters, however, require data in a long time window, which results in the time delay and affects the real-time control. It is not appropriate to increase accuracy with sacrificing the real-time capability of environmental classification.

In order to increase the accuracy and decrease the required time delay simultaneously for environmental classification, we construct a sequential model in this paper (\autoref{fig:1-Overview}). We hypothesize that we can fuse the sequential decisions (environmental classification results) from each frame of the image and decrease the required size of the time window by constructing a hidden Markov model (HMM) based on the probability theory and designing a decision fusion method. We verify our hypothesis through indoor and outdoor experiments with able-bodied subjects and amputees. The main contributions of this study include 1) constructing a hidden Markov model to optimize the sequential decisions of environmental classification, 2) designing a transition probability matrix for switching locomotion modes, 3) decreasing the time delay and increasing the accuracy of environmental classification simultaneously.

We organize the rest of our paper as follows. \autoref{sec:Methods} describes sequential decision fusion methods for environmental classification. Experimental results of presented methods are shown in \autoref{sec:Results}. After showing the results, we provide corresponding discussions in \autoref{sec:Discussion}. Finally, \autoref{sec:Conclusion} discusses the conclusion of this paper.

\section{Methods}\label{sec:Methods}
We present our decision fusion method in this section. We first state the research problems of environmental classification. To solve these problems, we briefly describe the methods of environmental feature extraction and classification based on the single image, which are introduced thoroughly in our previous paper \cite{zhang_environmental_2019}. Then we discuss how to construct a stable sequential model and fuse sequential decisions for environmental classification.

\subsection{Problem statement}
\label{subsec:problem}
Our objective is to classify environments accurately with short time delay. We can classify the current environment into a possible category. To determine a possible category, we need to calculate the probability distribution $\bm{Pr}$ of the current environment on each category first:
\begin{equation}\label{eq:Pr}
    \bm{Pr}(E = j) = p_j,\text{ }p_j > 0,\text{ }j = 1,..., 5, \text{ }\sum^5_{j=1}p_j = 1 
\end{equation}
where $p_1, ..., p_5$ represent the probability of that current environment $E$ is level ground ($j = 1$), up stairs ($j = 2$), down stairs ($j = 3$), up ramp ($j = 4$), or down ramp ($j = 5$).

We only use a depth camera to perceive environments, and thus the input data of our method are a series of depth images:
\begin{equation}
    \{x_k|k = 1, 2,..., t,..., t+d\}
\end{equation}
where $x_k$ is the image at time $k$. We denote the current time and delayed time as $t$ and $d$, respectively. 

To calculate the probability distribution $\bm{Pr_t}$ at current time $t$, we need to find a function $\bm{f_c}$ to classify the single image and a function $\bm{f_f}$ to fuse sequential decisions from different images:
\begin{equation}
    \bm{Pr_t}= \bm{f_f}\big(\bm{f_c}(x_1),..., \bm{f_c}(x_t),..., \bm{f_c}(x_{t+d})\big)
\end{equation}

There are some design constraints for the function of $\bm{f_c}$ and $\bm{f_f}$:
\begin{itemize}
    \item The classification function $\bm{f_c}$ should classify the single image accurately and quickly.
    \item The fusion function $\bm{f_f}$ should consider the relationship between the adjacent decisions.
    \item There might be some error images $x_k$, and thus $\bm{f_f}$ should tolerate some error decisions.
    \item The environmental classification accuracy should be high.
    \item The delayed time $d$ should be short.
\end{itemize}

\subsection{Preprocessing environmental images}
\begin{figure}[htbp]
    \centering
    \includegraphics[width=8cm]{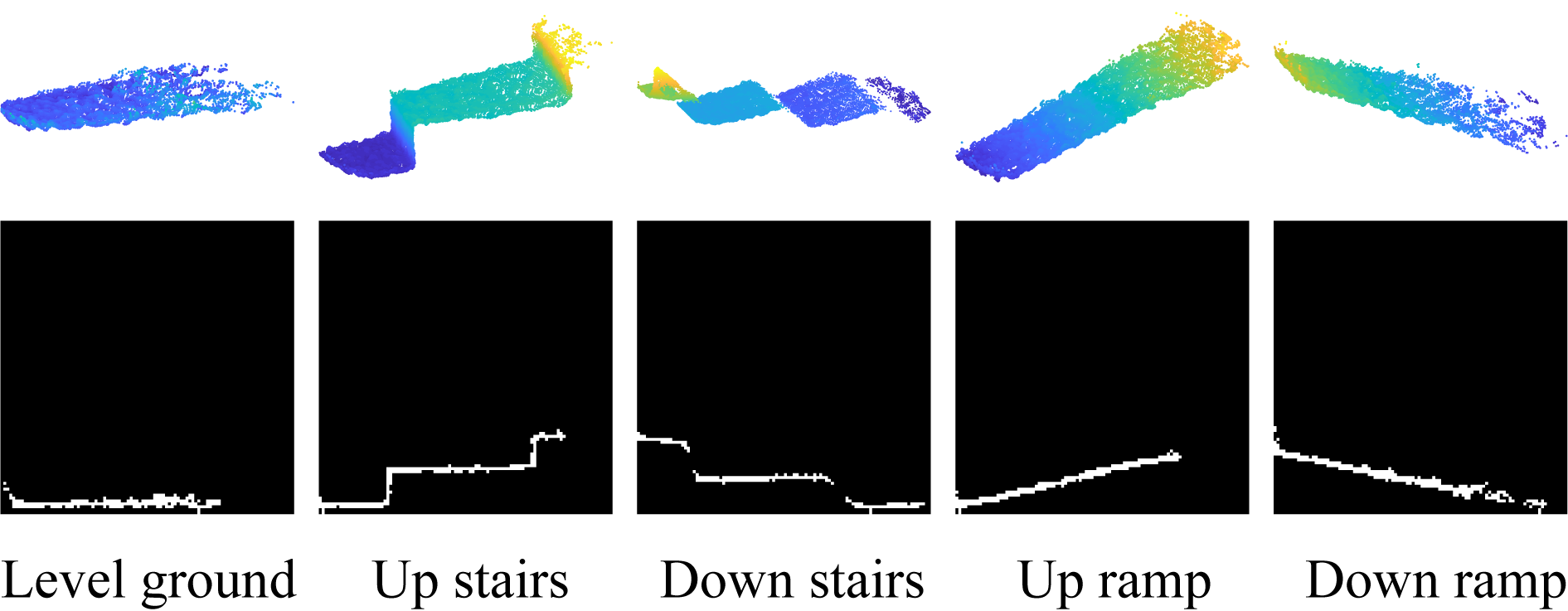}
    \caption{Environmental features extraction process. The point cloud on the first row is a set of 3D points. We convert the point cloud to the binary image, which only has two types of values: one or zero.}
    \label{fig:2-FeaturesExtraction}   
\end{figure}

The depth camera can output the point cloud of the environment, which is a set of three dimensional (3D) points in the space $\mathbb{R}^3$. A problem for point clouds is that they are unstable because the camera is worn on the leg and rotates together with the leg. To solve this problem, we offset the point cloud in real time using the measured angle from an IMU. Another problem is that point clouds are unstructured and unordered. To convert the point cloud to structured and ordered data, we project the point cloud to binary images (\autoref{fig:2-FeaturesExtraction}), which can be classified by the convolutional neural network (CNN) easily. The detail methods of offsetting and projecting point cloud are described in \cite{zhang_environmental_2019}. 

\subsection{Classifying the single environmental image}
After preprocessing environmental images, we need to find a classification function $\bm{f_c}$ to classify the single environmental image accurately and efficiently. We tend to select a suitable deep learning method to classify our environmental images because deep learning avoids designing features manually and has achieved great success in classifying images. Considering that our method should be efficient, we choose the convolutional neural network (CNN) as our classification function ($\bm{f_c}$ = CNN). CNN is efficient because it shares the weight parameters of the convolutional kernel and downsamples the image through max-pooling layers. 

\begin{figure*}[!h]
    \centering
    \includegraphics[width = 16cm]{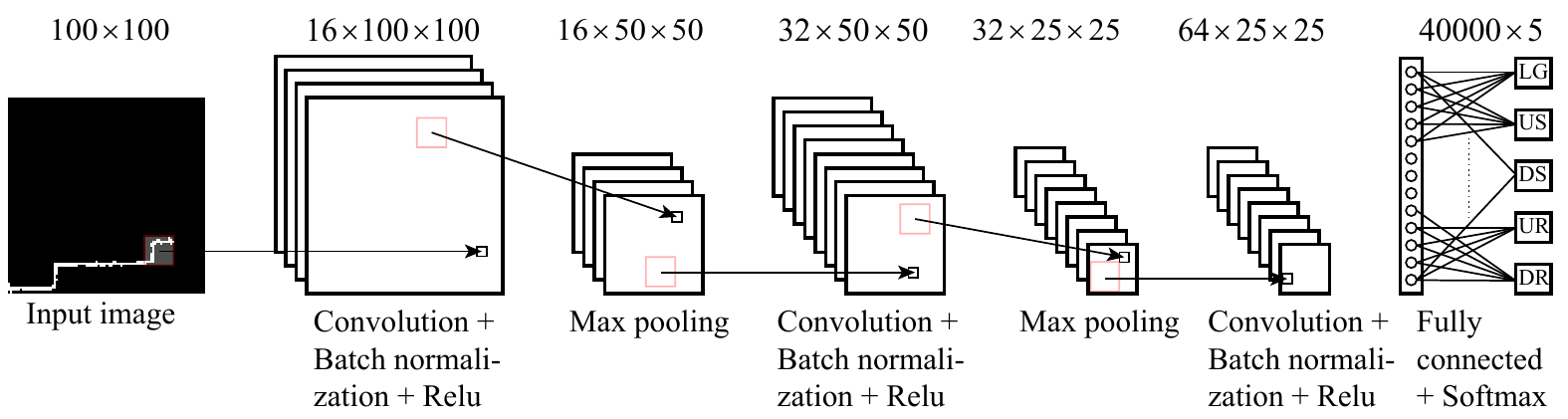}
    \caption{The architecture of the image classifier. The image classifier is based on a traditional CNN. The input and output of this CNN are a binary image of $100 \times 100$ pixels and classification scores on five types of environments. The definitions of LG, US, DS, UR, and DR are the same as in \autoref{fig:1-Overview}. We apply batch normalization and Relu activation function after each convolutional layer. The convolutional layer uses a filter to sum the bias and the dot product of pixels and corresponding parameters in the filter. Each input channel is normalized by the batch normalization. The Relu activation set the elements less than zero to zero. The max-pooling layers downsample the image and extract features in different resolutions. The output features from the last max-pooling layer are flattened and input to the fully connected layers. The fully connected layer calculates corresponding classification scores by adding the bias with the product sum of weights and features.}
    \label{fig:3-CnnModel}   
\end{figure*}

We then design an image classifier based on a simplified CNN \cite{krizhevsky_imagenet_2012}, which is shown in \autoref{fig:3-CnnModel}. The input of our classifier is a binary image of $100 \times 100$ pixels, and the output is the probability distribution (classification scores) of the current image on five types of categories. There are three convolutional layers and two max-pooling layers. The kernel size of the all convolutional layers is $3 \times 3$ pixels. As for the max-pooling layers, the kernel size is set to $2 \times 2$ pixels. There are 16, 32, and 64 channels for three convolutional layers. Moreover, We use batch normalization and Relu activation after each convolutional layer. 

Before training the network, we initialize all parameters randomly. The initial weight values of convolutional layers and fully connected layer are generated from a Gaussian distribution randomly. The mean and standard deviation of the Gaussian distribution are 0 and 0.01, respectively.

\subsection{Sequential model of environmental classification}
Our image classifier can generate a decision for each input image, and we need to fuse sequential decisions because classification result based on a single image is not robust. For instance, the camera may capture error images when the leg swings quickly. An intuitive method to fuse sequential decisions is to consider their temporal relationships. We can regard the captured images as sequence signals because human walking is continuous. After constructing the sequential model, we can design a hidden Markov model ($\bm{f_f}$ = HMM) to describe relationships between different decisions. 

\begin{figure*}[!h]
    \centering
    \includegraphics[width=16cm]{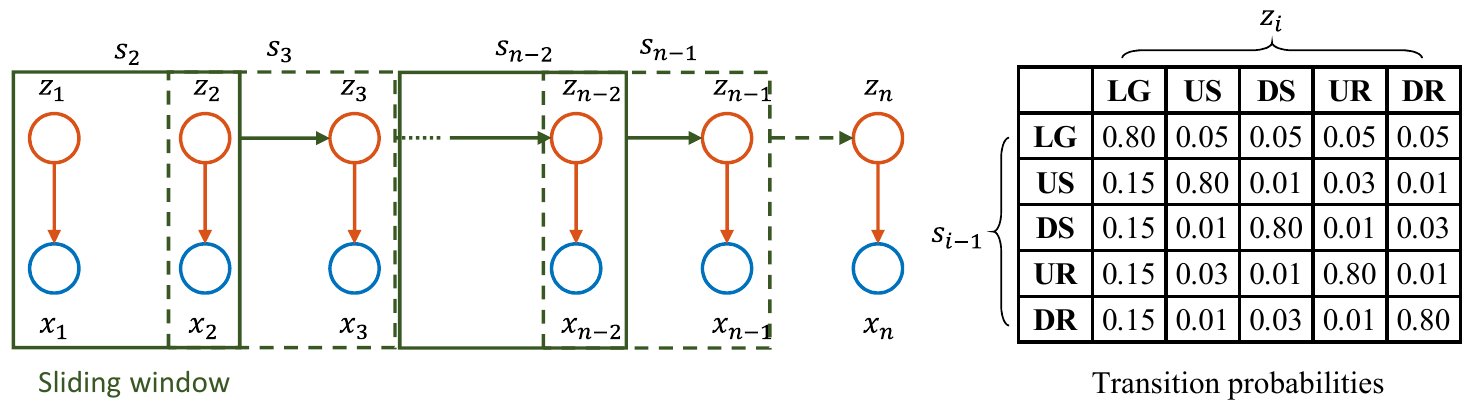}
    \caption{The hidden Markov model (HMM) of environmental classification. Latent states $z_i$ represent current environments, including level ground (LG), up stairs (US), down stairs (DS), up ramp (UR), and down ramp (DR). The observation is the current image, which is denoted by $x_i$. The $s_i$ is the smooth latent state that is calculated based on the mean value of previous latent states in a sliding window. The transition probabilities display the prior probabilities to transit from last smooth state $s_i$ to current state $z_i$. }
    \label{fig:4-HmmModel}   
\end{figure*}

There are two important elements in HMM: latent states and observations. As shown in \autoref{fig:4-HmmModel}, we can regard the current category and captured image of current environment as the latent state $z_t$ and observation $x_t$, respectively. The emission (conditional) probability $p(x_t|z_t)$ of observing image $x_t$ given the latent state $z_t$ is:
\begin{equation}
    \begin{split}
    &p(x_t|z_t = j) = \bm{f_c}(x_t)[j],\text{ }p(x_t|z_t = j) > 0,\text{ }\\ 
    &j = 1,..., 5,\text{ }\sum^5_{j = 1} p(x_t|z_t = j) = 1    
    \end{split}
\end{equation}
where $\bm{f_c}(x_t)$ denotes a vector of classification scores based on the current image $x_t$. The jth value of this vector equals to the probability of that the category of current environment is $j$. The definition of $j$ is the same as in (\ref{eq:Pr}).

The estimated category of $z_t$ may not be robust because there are some error images. In order to make the fusion function $\bm{f_f}$ to tolerate errors, we need to calculate a smooth state $s_t$ to substitute $z_t$. A simple method is to calculate the average probability distribution of the state $s_t$ in a sliding window:
\begin{equation}
    \begin{split}
    p(s_t = j) = 
    \begin{cases}
    p(x_t|z_t = j), &t \leq l_w\\
    \sum^t_{k=t-l_w}p(s_k = j)/l_w, &t > l_w
    \end{cases}   
    \end{split}
\end{equation}
where length of the sliding window is denoted as $l_w$.

In the HMM, adjacent latent states are connected by the transition probability, which represents the probability of transiting from a type of state to another type of state. The transition probability can be estimated base on our experience of life. For instance, the transition between different types of environments happens much less frequently than remaining in the same environment. Moreover, the stairs and ramp are usually connected by the level ground. Hence, we propose several rules to design the transition probability matrix $T$. We use $\bm{T}$ and $t_{ij}$ to denote transition probability matrix and the transition probability from $i$ to $j$, respectively:
\begin{itemize}
    \item The probabilities of remaining the same environment ($t_{ii}$) are higher than that of transiting to different environments ($t_{ij},~i \neq j $).
    \item The probabilities of remaining the same environment ($t_{ii},~i = 1,..., 5$) are same for all environments.
    \item The probabilities of transiting from level ground to other types of environments ($t_{1j},~j = 2,..., 5$) are same.
    \item The probabilities of transiting from other types of environments to the level ground ($t_{i1},~i = 2,..., 5$) are same.
    \item The probabilities of transiting between different upward environments or downward environments ($t_{ij},~i \neq j,~i = 2,..., 5,~j = 2,..., 5,~|i - j|\text{ is even}$) are same and low.
    \item The probabilities of transiting between the upward environments and the downward environments ($t_{ij},~i \neq j,~i = 2,..., 5,~j = 2,..., 5,~|i - j|\text{ is odd}$) are same and the lowest.
\end{itemize}

According to the above rules, we design a transition probability matrix, which is shown in \autoref{fig:4-HmmModel}.

\subsection{Sequential decision fusion}
Instead of using voting strategy or median strategy \cite{krausz_depth_2015, zhang_environmental_2019}, we modify the Viterbi algorithm to fuse the probability distribution of sequential decisions and estimate the smooth state $s_i$ \cite{viterbi_error_1967}. The voting and median strategies are not appropriate because they do not consider the credibility of different decisions. However, the decision whose probability distribution concentrates on one category is more credible than those whose probabilities distributed similarly on all categories. 

Considering that the credibilities of decisions are different, our modified Viterbi algorithm takes account of the probability distribution of every decision (Algorithm \ref{algo:Viterbi}). Our method is able to tolerate some errors because the decisions from error images are usually less credible than the stable decisions. After using our method, we can classify environments accurately with only delaying one frame.

However, if there are many error images, we still need a voting strategy to increase the robustness of the classification results further. The voting strategy is to calculate the mode of a series of smooth states in a sliding window:
\begin{equation}
    d_v = \text{mode}\{s_k| k = t - l_v,..., t + l_v\}
\end{equation}
where $d_v$ is the final decision of voting strategy and $l_v$ is the number of delayed frames caused by the voting strategy. 

Consequently, the decision fusion function can be our HMM ($\bm{f_f}$ = HMM) or the combination of our HMM and voting strategy ($\bm{f_f}$ = HMM + Voting). The symbol of $+$ denotes combination.

\begin{algorithm*}[!h]
\SetAlgoLined
\textbf{Input:} Emission probability $\bm{p}(x_t|z_t)$ at time $t$, transition probability $t_{ij}$, and smooth window length $l_w$.

\textbf{Output:} Optimized smooth state of the last time $s^{\text{opt}}_{t-1}$ and the probability distribution of the current smooth state $\bm{p}(s_t)$.

\textbf{Initialize:} $\bm{p}(s_t) = \bm{p}(x_t|z_t)$, $t \leq l_w$.

\textbf{Repeat:}
\begin{itemize}
    \item Calculate the average probability distribution in a sliding window:
    $\overline{\bm{p}}(s_{t-1}) = \sum^{t-1}_{k = t - l_w} \bm{p}(s_k)/ l_w$
    \item Calculate the posterior probability distribution of the last smooth state:
    $\hat{\bm{p}}(s_{t-1}) = [\hat{p}(s_{t-1} = i) = \sum^5_{j = 1} \overline{p}(s_{t-1} = i)\cdot t_{ij} \cdot p(x_t|z_t = j) \text{ }|\text{ } i = 1, ..., 5]$
    
    \item Set the last smooth state as the smooth state with the maximum probability:
    $s^\text{opt}_{t-1} = \text{argmax} \big( \hat{\bm{p}}(s_{t-1})\big)$
    
    \item Update the posterior probability distribution of the current smooth state: 
    $\hat{\bm{p}}(s_t) = [\hat{p}(s_t = j) = \overline{p}(s_{t-1} = s^\text{opt}_{t-1})\cdot t_{ij} \cdot p(x_t|z_t = j) \text{ }|\text{ } j = 1, ..., 5]$
    
    \item Normalize the probability distribution of the current smooth state:
    $\bm{p}(s_t) = [p(s_t = j) = \hat{p}(s_t = j) / \sum^5_{i = 1} \hat{p}(s_t = i) \text{ }|\text{ } j = 1, ..., 5]$

\end{itemize}
\caption{Optimizing smooth states for environmental classification.}
\label{algo:Viterbi}
\end{algorithm*}

\subsection{Experimental setup}
We evaluated our method by inviting subjects to implement indoor and outdoor experiments, which are the same as in \cite{zhang_environmental_2019}. We invited able-bodied subjects and amputees to wear our sub-vision system above the knee joint to capture environmental images. The sub-vision system consists of a depth camera (CamBoard pico flexx, $68 \times 7 \times 25\;mm$, pmdtechnologies) and an IMU (MTi 1-series, $12.1 \times 12.1 \times 2.55\;mm$,  Xsens Technologies). During the experiments, we requested each subject to walk in an experimental area repeatedly for five times. In each trial, there are three level ground modes, one up and one down stairs modes, and one up and one down ramp modes. 

We used the trained CNN model in our previous research to test our decision fusion methods. The detailed training settings of the CNN model are shown in \cite{zhang_environmental_2019}. We utilized the trained CNN model to calculate the original classification scores (emission probabilities $\bm{p}(x_k|z_k)$) from collected images. Then we implemented our decision fusion methods to estimate the final decisions ($d_v$), which were compared with the actual modes to evaluate the classification accuracy of our method.

We implemented the experimental analysis on a computer with an Intel Core i7-7700K (4.2 GHz), a 16 GB DDR3, and a graphics card (NVIDIA GeForce GTX 1050 Ti). The program is based on MATLAB@ R2017b.

\subsection{Statistical analysis}
In our experiments, we collected the generated binary images and labeled the actual modes manually. The mean and standard deviations of classification accuracy were analyzed for different subjects. We utilized a t-test at a significance level of $\alpha = 0.05$ and $P$ value to evaluate the significance of the difference between the results using different methods. The $P$ value is the probability that the null hypothesis is true.

\section{Results}\label{sec:Results}
\subsection{Subject information}
Five able-bodied subjects and three transfemoral amputees participated in our experiments. We provide the basic information of subjects and amputees in \autoref{tab:subjectInfo} and \autoref{tab:amputeeInfo}. We recruited the amputees from a local prosthetic company. The able-bodied subjects are from our university. One of the able-bodied subjects is the author of this paper. The approval to perform these experiments was granted by the Review Board of Southern University of Science and Technology. Subjects and amputees signed informed consents before the experiments.

\begin{table}[htbp]
\centering
\caption {\label{tab:subjectInfo} Basic information of able-bodied subjects.}
\renewcommand{\arraystretch}{1.5} 
\begin{center}
\begin{tabular}{l c c c c}
\toprule
Subjects & Height (m) & Weight (kg) & Age (years) & Gender \\
\midrule
Subject 1 & 1.66 & 59 & 28 & Male \\
Subject 2 & 1.65 & 63 & 30 & Male \\
Subject 3 & 1.68 & 58 & 29 & Male \\
Subject 4 & 1.72 & 60 & 24 & Female \\
Subject 5 & 1.67 & 53 & 25 & Male 
\end{tabular}
\end{center}
\end{table}

\begin{table}[htbp]
\centering
\caption {\label{tab:amputeeInfo} Basic information of amputees.}
\renewcommand{\arraystretch}{1.5} 
\begin{center}
\begin{tabular}{l c c c}
\toprule
Subjects & Amputee 1 & Amputee 2 & Amputee 3\\
\midrule
Height (m) & 1.70 & 1.70 & 1.69 \\
Weight (kg) & 64 & 60 & 62\\
Age (years) & 38 & 38 & 42 \\
Gender & Male & Male & Male \\
Amputation time & 2016 & 2001 & 2000\\
Amputation side & Left & Right & Left \\
Residual limb length (m) & 0.33 & 0.30 & 0.31
\end{tabular}
\end{center}
\end{table}

\subsection{Environmental classification results}
\begin{figure}[htbp]
    \centering
    \includegraphics[width=8cm]{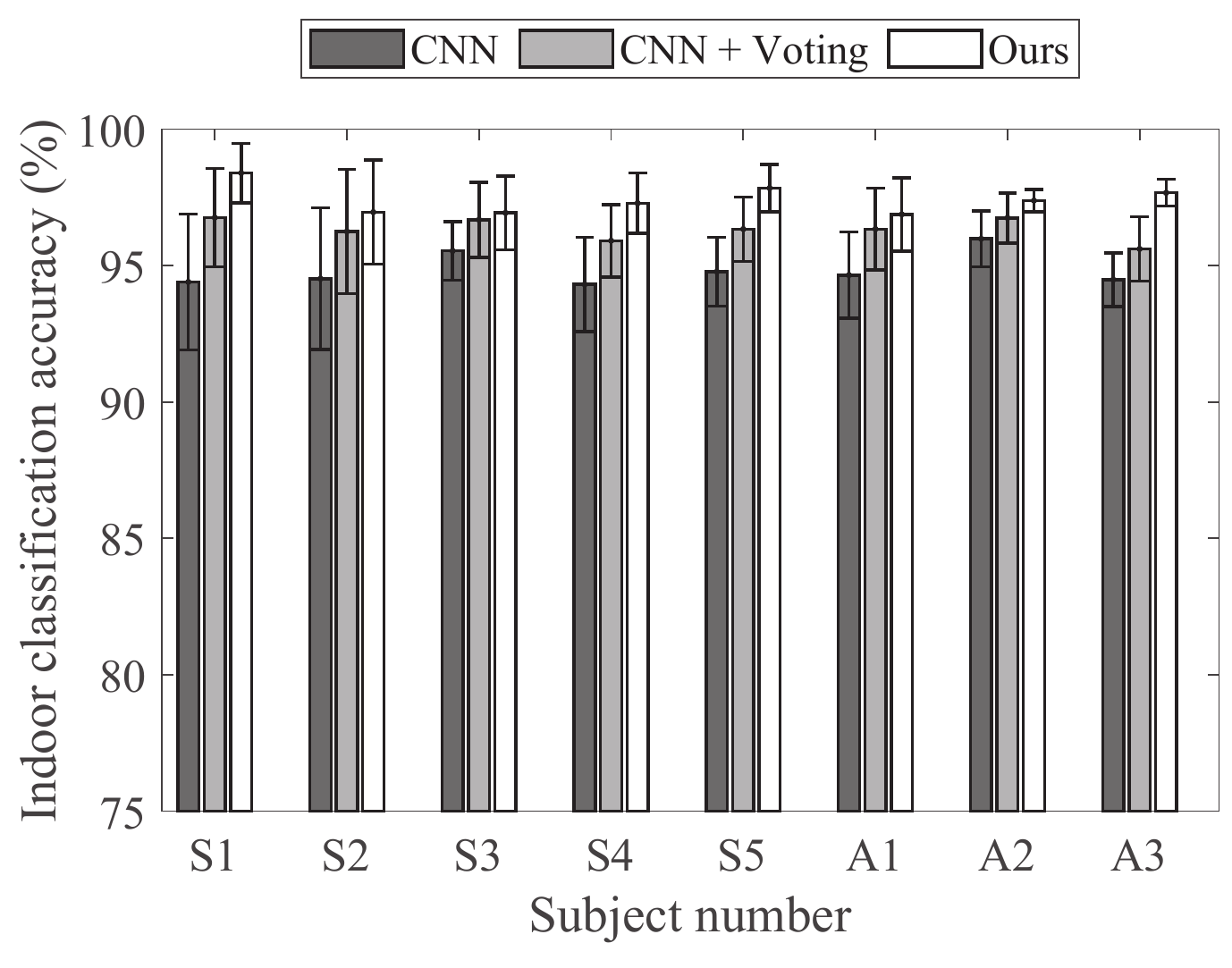}
    \caption{Comparison of indoor environmental classification results. Subjects include five able-bodied subjects (S1-S5) and three transfemoral amputees (A1-A3). The error bars represent mean $\pm$ one standard deviation of classification accuracy in five repeated experiments.}
    \label{fig:5-ClassificationResultsIndoor}   
\end{figure}

\begin{figure}[htbp]
    \centering
    \includegraphics[width=8cm]{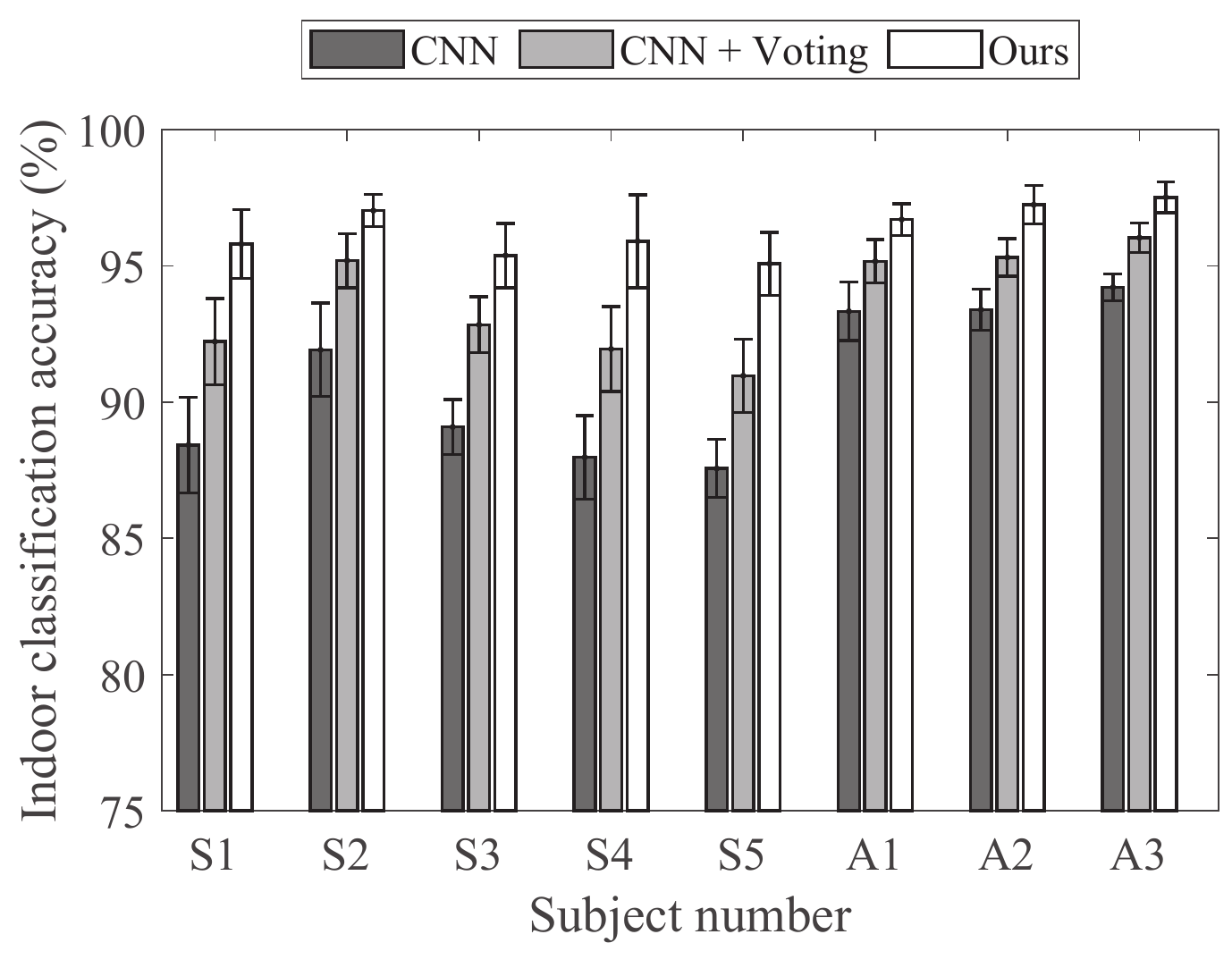}
    \caption{Comparison of outdoor environmental classification results. The meaning of the error bar and subject number are the same as in \autoref{fig:5-ClassificationResultsIndoor}.}
    \label{fig:6-ClassificationResultsOutdoor} 
\end{figure}

In order to evaluate the performance of our method, we compared the environmental results of our method ($\bm{f_c} + \bm{f_f}$ = CNN + HMM + Voting) with that of CNN and the combination of CNN and voting strategy (CNN + Voting). We set the length of sliding window $l_w$ and the number of delayed frames $l_v$ at 5 and 1, respectively. Then we calculated the mean and standard deviation (SD) of the indoor and outdoor environmental classification accuracy using the above three methods. The error bars and statistical data of classification accuracy are shown in \autoref{fig:5-ClassificationResultsIndoor}, \autoref{fig:6-ClassificationResultsOutdoor}, and \autoref{tab:classification}.

The classification accuracy using our method is statistically different from that using (CNN + Voting) ($P < 0.001$). As shown in \autoref{tab:classification}, Compared to the (CNN + Voting), our method increases the mean values of classification accuracy in the indoor experiment and outdoor experiment by 1.09\% and 2.62\%, respectively. Moreover, the standard deviations of classification accuracy decrease after using our method. Hence, our method can classify environments accurately and stably.
 
Moreover, we compare the classification accuracy for each subject using three different methods. As shown in \autoref{fig:5-ClassificationResultsIndoor} and \autoref{fig:6-ClassificationResultsOutdoor}, our method can increase the classification accuracy for all subjects and amputees in the indoor and outdoor experiments. 

\begin{table}[htbp]
\centering
\caption {\label{tab:classification} Comparison of environmental classification accuracy. The number of delayed frames is one.}
\renewcommand{\arraystretch}{1.5} 
\begin{center}
\begin{tabular}{l c c c c}
\toprule
Methods & Mean (\%) & SD (\%) & Mean (\%) & SD (\%) \\
\midrule
 & \multicolumn{2}{c}{Indoor} & \multicolumn{2}{c}{Outdoor}\\
\midrule
CNN & 94.83 & 1.64 & 90.74 & 2.84 \\
CNN + Voting & 96.33 & 1.41 & 93.71 & 2.09 \\
\textbf{Ours}& \textbf{97.42} & \textbf{1.17} & \textbf{96.33} & \textbf{1.28}
\end{tabular}
\end{center}
\end{table}

\subsection{Trade-off between the accuracy and time delay}
We can increase the classification accuracy further by using the voting strategy because human remains the same locomotion modes in most situations. The voting strategy, however, can cause the time delay. Here we analyze the trade-off between classification accuracy and time delay. 

\begin{figure}[htbp]
    \centering
    \includegraphics[width=8cm]{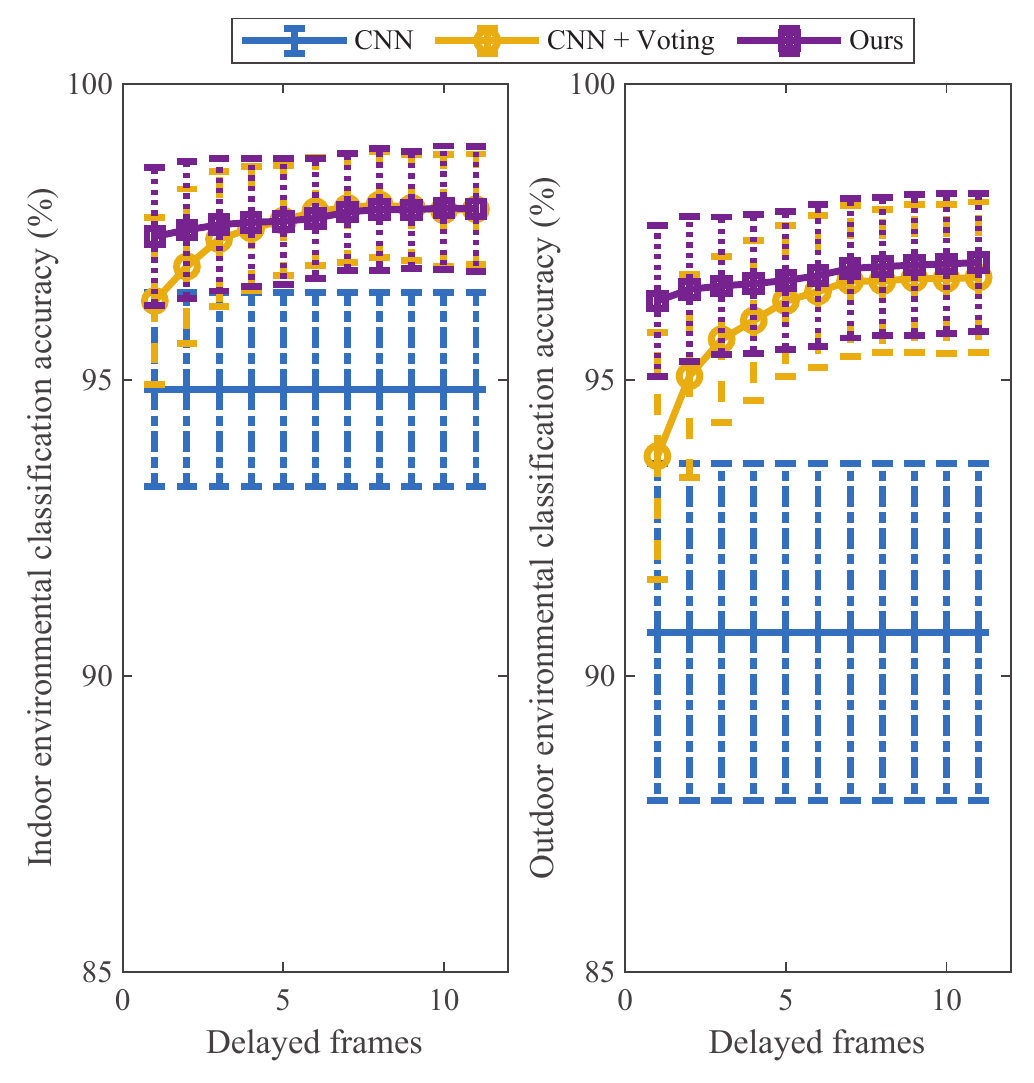}
    \caption{The relationship between the classification accuracy and the number of delayed frames. The definitions of error bars and different methods are the same as in \autoref{fig:5-ClassificationResultsIndoor}.}
    \label{fig:7_time_delay_analysis}   
\end{figure}

We calculated environmental classification accuracy using three different methods and different window length $w_v$ of voting strategy. The number of delayed frames $l_v$ for (CNN + Voting) equals to $(w_v-1)/2$. Meanwhile, $l_v$ for our method is $(w_v+1)/2$ because our HMM can also cause one frame delay. We aligned the classification accuracy of three different methods based on the number of delayed frames, which is shown in \autoref{fig:7_time_delay_analysis}. The classification accuracy of our method and (CNN + Voting) increases with the number of delayed frames. Our method is affected less by the number of delayed frames than (CNN + Voting). We calculated the slope of classification accuracy relative to the number of delayed frames. In the indoor experiments, the mean slope of our method and (CNN + Voting) are 
0.047\% and 0.15\%, respectively. In the outdoor experiments, the above two values change to 0.065\% and 0.30\%. 

Moreover, we analyzed the difference of time delay between using our method and using (CNN + Voting) to achieve the same classification accuracy (difference of accuracy $<$ 0.05\%). In the indoor experiments, the classification accuracy of our method achieves 97.53\% with a delay of two frames. Meanwhile, (CNN + Voting) requires a delay of four frames to achieve the accuracy of 97.56\%. In the outdoor experiments, the classification accuracy of our method with a delay of two frames is 96.53\%. (CNN + Voting) achieves 96.49\% with the delay of 6 frames. Considering that the capturing frequency of our depth camera is 15 frames per second, our method can decrease the time delay by about 133 ms and 267 ms to achieve the same classification accuracy as (CNN + Voting).

\subsection{Sequential decisions of environmental classification}
\begin{figure}[htbp]
    \centering
    \includegraphics[width=8cm]{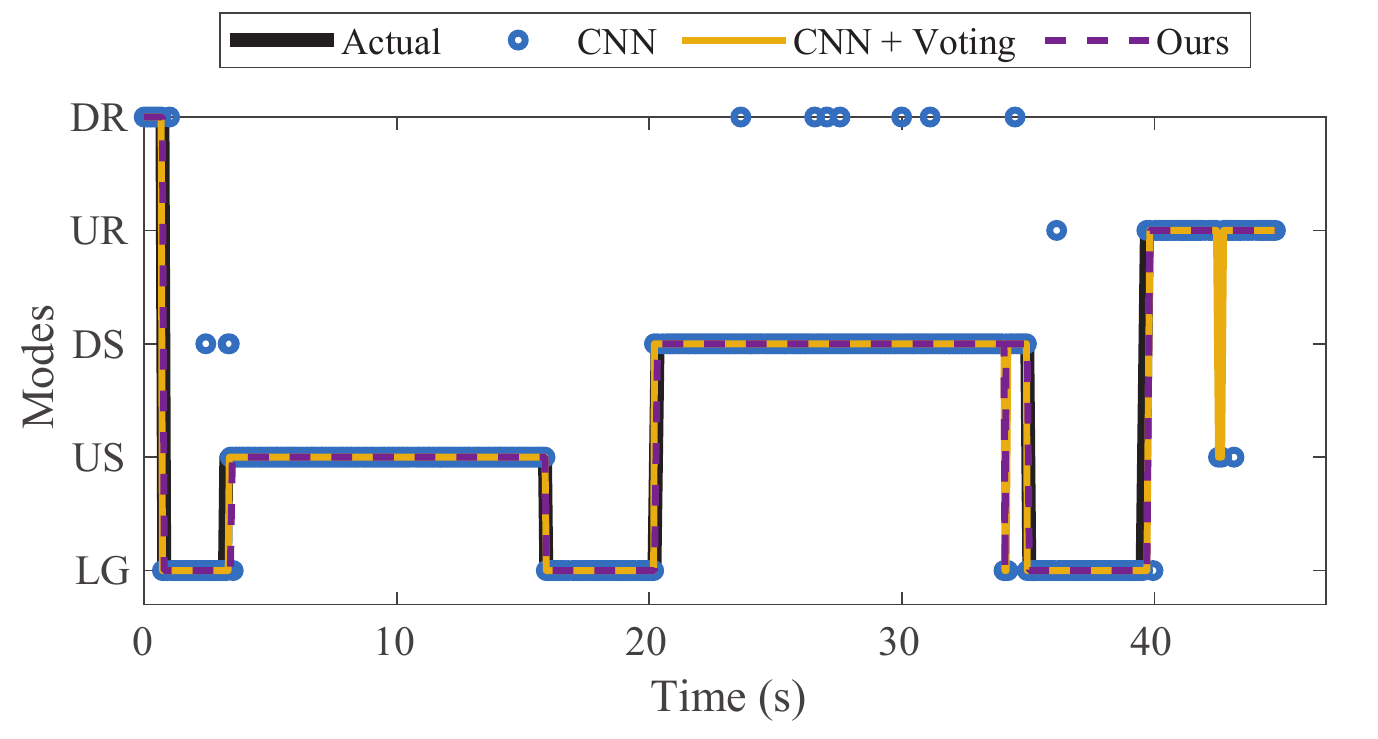}
    \caption{Sequential decisions of environmental classification in the indoor experiment. Amputee 1 is the subject of this experiment. The black heavy line denotes the actual modes, which are labeled manually based on the captured binary images. The blue circles, yellow thin line, and purple dash line represent the classification modes using CNN, (CNN + Voting), and our method. LG, US, DS, UR, and DR are the abbreviations of level ground, up stairs, down stairs, up ramp, and down ramp.}
    \label{fig:8-DecisionModesComparisonIndoor}   
\end{figure}

\begin{figure}[htbp]
    \centering
    \includegraphics[width=8cm]{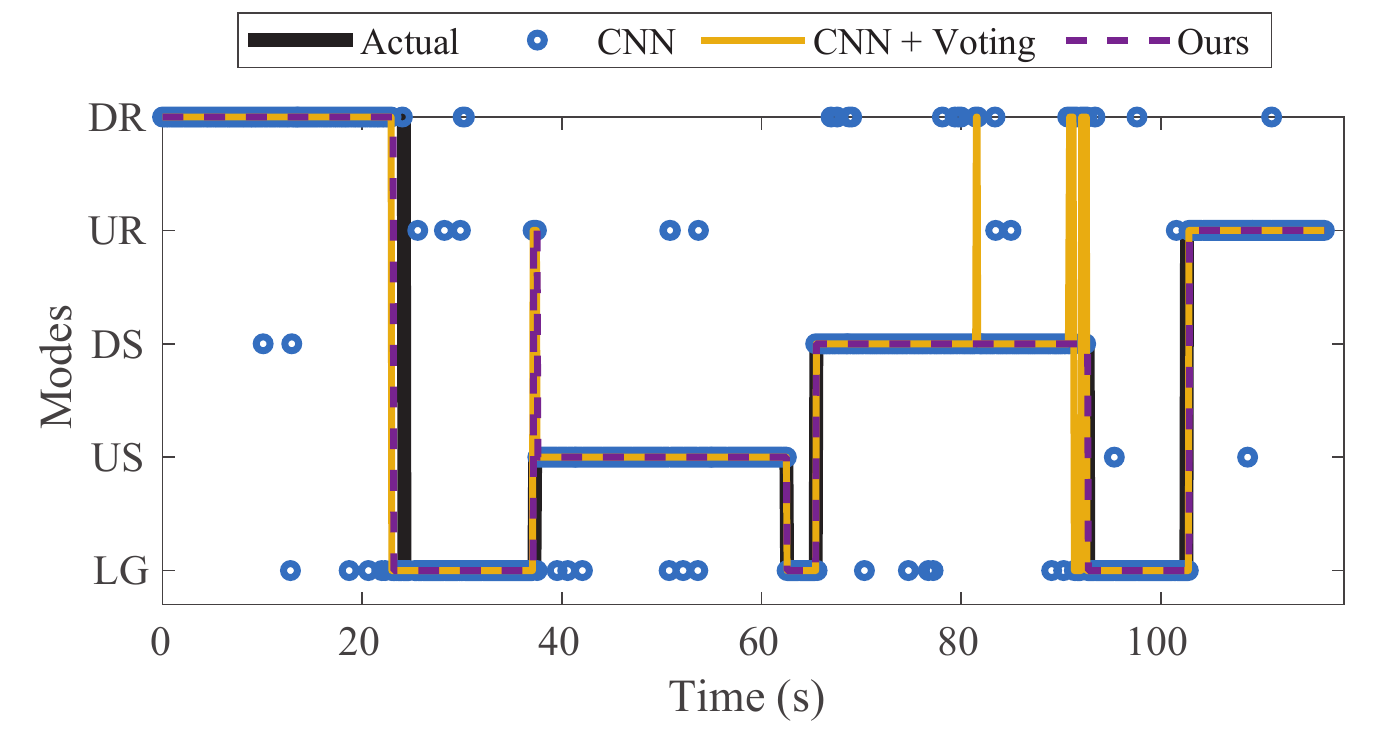}
    \caption{Sequential decisions of environmental classification in the outdoor experiment. Amputee 2 is the subject in this experiment. Definitions of the legend and labels are the same as in \autoref{fig:8-DecisionModesComparisonIndoor}.}
    \label{fig:9-DecisionModesComparisonOutdoor}   
\end{figure}

We visualize the sequential decisions of environmental classification intuitively in \autoref{fig:8-DecisionModesComparisonIndoor} and \autoref{fig:9-DecisionModesComparisonOutdoor}. The original classification results of CNN are noisy, which are not suitable to control the prosthesis. Then we utilized the voting strategy to filter the classification results and set the number of delayed frames $l_v$ to 2. The classification results using (CNN + Voting) become clearer but still have some error results. After using our method, the error classification results only happen once in both the indoor and outdoor experiments. Consequently, under the same number of delayed frames, our method can improve the classification results more than (CNN + Voting).

\subsection{Comparison of probability distributions}
\begin{figure}[htbp]
    \centering
    \includegraphics[width=8cm]{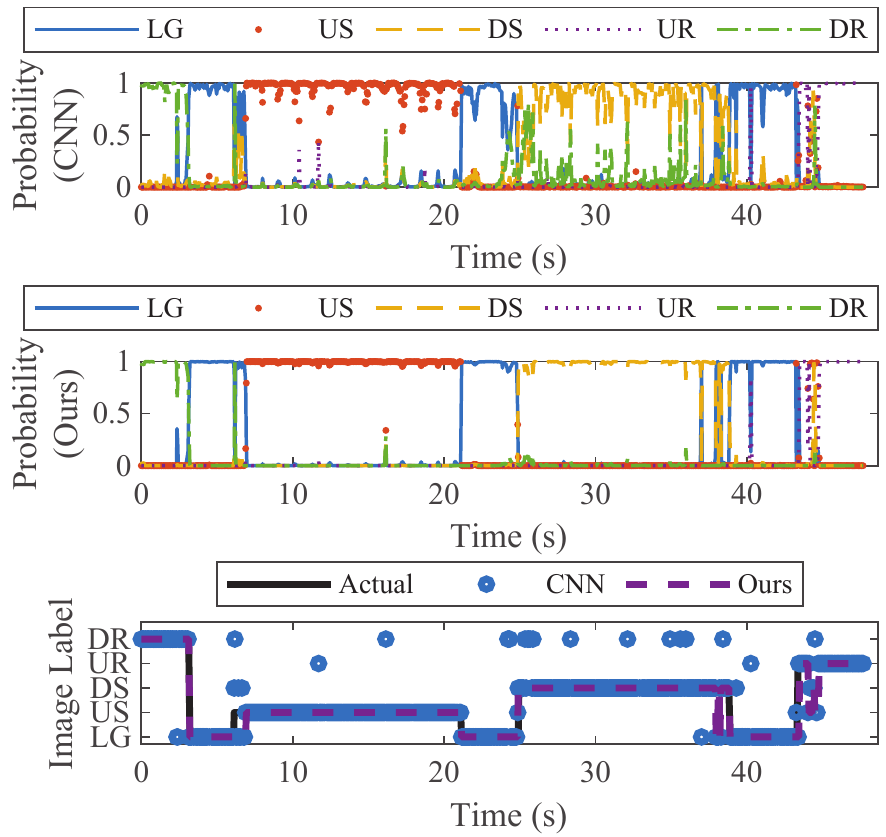}
    \caption{Probability distributions calculated by different methods. In two upper figures, the blue solid line, red dot, yellow dash line, purple dot line, and green dot dash line represent level ground (LG), up stairs (US), down stairs (DS), up ramp (UR), and down ramp (DR).} 
    \label{fig:10-ProbabilityVariation}   
\end{figure}

Our method is better than CNN and (CNN + Voting) because we consider the probability distribution in each frame and the transition probability between adjacent states. As shown in \autoref{fig:10-ProbabilityVariation}, the original probability distribution calculated by the CNN varies at different frames. There are also some error probability distributions in the original results because of the intense rotation of cameras or some anomalous environments. The posterior probability distributions of our method are more discernable than that of CNN. Most probability distributions concentrate on one mode.

\section{Discussion}\label{sec:Discussion}
\subsection{Advantages of our method}

In this research, we proposed a concise method to fuse sequential decisions to increase environmental classification accuracy and decrease the time delay. Compared to the traditional voting strategy \cite{krausz_depth_2015, zhang_environmental_2019}, our methods have several advantages.

Firstly, our method takes account of the credibility of each decision. For traditional voting strategy, all decisions contribute equally, but this is not credible. In real situations, the camera may provide error images sometimes. For instance, the camera cannot perceive front terrains when its orientation angle in the sagittal plane is too big or too small. Also, there are some interference objects, including uneven ground and curbs, in the environments, especially in the outdoor environment. The probability distributions of these error decisions are more ambiguous than that of normal decisions. We can decrease the credibility of error decisions after fusing the probabilities, which is better than traditional voting strategy.

Additionally, we designed a transition probability matrix base on the characteristics of human walking and daily environments. This transition probability matrix provides the relationship between the last state and current state, and thus optimizes the accuracy of the environmental classification. As stated in \autoref{sec:Results}, our method increases classification accuracy of all subjects in both indoor and outdoor environments. 

Moreover, our method is suitable for real-time control because it has low computational complexity and requires a short time delay. It only takes 0.02 ms to use our method to update one decision. Besides, our method can still achieve high accuracy with a delay of only one frame. As shown in \autoref{fig:7_time_delay_analysis}, our method is less affected by the number of delayed frames than the method of (CNN + Voting). In our previous research \cite{zhang_environmental_2019}, we also achieved high classification accuracy, but we sacrificed the real-time performance. Although the camera can perceive the environments in front of the prosthesis and can tolerate time delay of recognition, large time delay decreases the response speed of the whole control system and cannot handle unexpected situations. Compared to traditional voting strategy, our method can achieve the same classification accuracy with decreasing time delay of 133 ms and 266 ms in the indoor and outdoor environments, respectively. Consequently, our method can increase the response speed of the control system.

Furthermore, we can also apply our method to classify human intent. The input of our decision fusion method is only the probability distribution. Hence our method is not limited in the environmental classification. The change of sensors does not affect our decision fusion method. Some human signals, such as EMG and IMU, can also be utilized to classify human motion intent during walking in complex environments. The requirements of real-time performance for these human signals are higher than visual signals because these human signals generate only tens of milliseconds ahead of motion or even after the motion. Then we need to achieve high classification accuracy with short time delay. As stated before, the required time delay of our method is down to one frame, and our method has low computational complexity. Thus, our method fulfills the above requirements.

\subsection{Limitations and future works}
Although our method can classify environments accurately and with short time delay, there are some limitations. Firstly, we have not applied our method on the real-time control of a powered prosthesis. The situations in real-time control can be different from that in the offline analysis. Besides, the amputees wearing powered prostheses may walk differently from that wearing passive prostheses. Hence, we will apply our method on the real-time control of the powered prosthesis to evaluate the performance of our method further. Moreover, the environmental classification can only provide prior information about human motion intent. Therefore, we need to fuse the decisions from visual signals and that from human signals to estimate human motion intent more accurately.

\section{Conclusion}\label{sec:Conclusion}
In this paper, we constructed a hidden Markov model and designed a transition probability matrix for environmental classification in assistive walking. We considered the probability distribution of original decision from the CNN and fused sequential decisions to increase the classification accuracy with short time delay. We invited able-bodied subjects and amputees to implement experiments in indoor and outdoor experiments. According to the experimental results, our method achieved the classification accuracy of 97.42\% and 96.33\% with delaying only one frame in the indoor and outdoor experiments, which were 1.09\% and 2.62\% higher than that using the traditional voting strategy. For achieving same classification accuracy, our method decreased time delay by 133 ms and 266 ms in the indoor and outdoor experiments compared to traditional voting strategy. Moreover, our decision fusion method only cost 0.02 ms to update one decision. Hence, our method realized our target: increasing the classification accuracy and decreasing the time delay simultaneously. These satisfactory experimental results validated the accuracy and real-time capability of our method, which is significant to improve the performance of prostheses.

\section*{Acknowledgment}
We acknowledge funding and support by the National Natural Science Foundation of China under Grant U1613206, 61533004, and 91648203, and in part by Guangdong Innovative and Entrepreneurial Research Team Program under Grant 2016ZT06G587.

\bibliographystyle{IEEEtran}
\bibliography{Decision_fusion}

\end{document}